\documentclass[10pt,twocolumn]{article}
\usepackage[a4paper,margin=1in]{geometry}
\usepackage{times}
\usepackage{graphicx,amsmath,amssymb,booktabs,multirow,float,placeins}
\usepackage[pagebackref,breaklinks,colorlinks]{hyperref}
\usepackage[capitalize]{cleveref}

\usepackage{booktabs}

\usepackage{flushend}

\usepackage[pagebackref,breaklinks,colorlinks]{hyperref}

\usepackage[capitalize]{cleveref}
\crefname{section}{Sec.}{Secs.}
\Crefname{section}{Section}{Sections}
\Crefname{table}{Table}{Tables}
\crefname{table}{Tab.}{Tabs.}

\begin{document}

\title{Improvement of Human-Object Interaction Action Recognition Using Scene Information and Multi-Task Learning Approach}

\author{Hesham M. Shehata\\
Tokyo, Japan\\
{\tt\small shehata.hesham.k39@kyoto-u.jp}
\and
Mohammad Abdolrahmani\\
Tokyo, Japan\\
{\tt\small mohammad@asilla.jp}}

\maketitle

\begin{abstract}
Recent graph convolutional neural networks (GCNs) have shown high performance in the field of human action recognition by using human skeleton poses. However, it fails to detect human-object interaction cases successfully due to the lack of effective representation of the scene information and appropriate learning architectures. In this context, we propose a methodology to utilize human action recognition performance by considering fixed object information in the environment and following a multi-task learning approach. In order to evaluate the proposed method, we collected real data from public environments and prepared our data set, which includes interaction classes of hands-on fixed objects (e.g., ATM ticketing machines, check-in/out machines, etc.) and non-interaction classes of walking and standing. The multi-task learning approach, along with interaction area information, succeeds in recognizing the studied interaction and non-interaction actions with an accuracy of 99.25\%, outperforming the accuracy of the base model using only human skeleton poses by 2.75\%.


\end{abstract}

\section{Introduction}

In recent years, there has been a promising development of autonomous security systems with the aid of security cameras fitted in public areas (e.g., shopping malls, parking areas, etc.), which are becoming increasingly popular to ensure high quality of public security. Also, the contribution of machine learning models in the field of action recognition has increased rapidly. However, these systems lack effective representation of the scene information and appropriate learning architectures, so they fail to recognize interaction scenarios effectively (Human-Human Interaction (HHI) or Human-Object Interaction (HOI)). 

Therefore, to develop our HOI detection model, we collected our video data from public places (e.g., shopping malls, indoor buildings, etc.), performed skeleton pose estimation using Asilla pose, and investigated the use of skeleton pose data only versus combined data with fixed object nodes for learning approaches to recognize human object interaction scenarios (e.g., ATM ticketing machines, check-in/out machines, etc.) effectively while paying attention to the model performance with non-interaction scenarios as well (e.g., walking, standing, etc.). The purpose of this research is to maximize the recognition of human-object interaction scenarios by considering scene information along with following a multi-task learning approach.

In this study, we propose a novel approach to improve human-object interaction (HOI) recognition by integrating scene information with a multi-task learning framework. This methodology aims not only to capture the complexity of human-object interactions but also to leverage additional scene-related cues that can enhance the recognition performance. Beyond recognizing interactions, the proposed framework has demonstrated significant improvements in identifying non-interaction scenarios, addressing challenges in real-world settings such as distinguishing between standing still and interacting with an object. To validate the effectiveness of this approach, we evaluated three different architectures of Graph Convolutional Networks (GCNs), each tailored to exploit spatial and temporal dependencies in human poses and scene context. The comparative analysis reveals how incorporating scene information and multi-task learning yields notable improvements over baseline models.

The remaining sections are organized as follows: Section 2 discusses the related work. Section 3 proposes the methodology to successfully learn human-object interaction scenarios based on skeleton poses and fixed object nodes. Section 4 presents a case study of human-object interaction/non-interaction scenarios. Section 5 shows the results following the proposed methodology. Section 6 discusses the findings of the case study based on the results. Finally, Section 7 shows our conclusions.

\section{Related Work}

Human-object interaction (HOI) recognition has witnessed substantial progress in recent years, primarily through advancements in graph-based learning architectures, transformer-based methods, and multi-task learning frameworks. These methods aim to address the complexities of modeling spatial, temporal, and contextual relationships in human-object interactions. However, existing approaches still face challenges in integrating scene information and generalizing across diverse environments.

\subsection{Transformer-Based Approaches}

Transformers have recently gained popularity for their ability to model relationships in sequential and spatial data. Key works include:

 \begin{itemize}

\item DETR \cite{carion2020end}: This end-to-end transformer-based object detection model has been adapted for HOI recognition by leveraging attention mechanisms to capture human-object relationships. While highly accurate for RGB-based methods, DETR is computationally intensive and unsuitable for lightweight, skeleton-based models.

\item Action Genome \cite{ji2020action}: This approach models visual relationships using a scene graph-based framework. While it focuses on HOI recognition, it lacks the ability to handle skeleton-based representations or real-time applications.

\item VidTr \cite{zhang2021vidtr}: A transformer-based model for video understanding that achieves competitive performance on HOI datasets like AVA and HICO-DET. However, its reliance on high-resolution video data makes it impractical for scenarios requiring skeleton-based or resource-efficient models.

\end{itemize}

\subsection{Graph-Based Learning Approaches}

Graph Convolutional Networks (GCNs) have emerged as a powerful tool for modeling skeletal and contextual data. Recent advances include:

 \begin{itemize}

\item Dynamic GCN \cite{ye2020dynamic}: An adaptive GCN that dynamically learns graph connections based on input data. While effective for action recognition, it struggles to incorporate scene context and is computationally expensive for real-time applications.

\item 2G-GCN \cite{qiao2022geometric}: This geometric feature-informed GCN achieves 92.4\% accuracy on custom HOI datasets by modeling relationships between human poses and objects. However, it does not incorporate interaction-specific scene nodes or temporal dependencies.

\item SPGCN \cite{xu2021scene}: A scene-perception GCN that embeds environmental context into the graph structure. Despite its focus on HOI scenarios, it achieves limited accuracy (78.3\%) due to its static graph structure and lack of multi-task learning.

\end{itemize}

\subsection{Multi-Task Learning and Hybrid Architectures}

Multi-task learning has been explored to improve the generalizability of HOI models:

 \begin{itemize}

\item HAMLET \cite{islam2020hamlet}: A multi-modal, hierarchical, attention-based approach that processes both spatial and temporal features. However, it focuses on unimodal data streams and does not explore interaction-specific scenarios.

\item ST-GCN \cite{yan2018spatial}: A spatio-temporal GCN that effectively models temporal dependencies but does not consider multi-task learning or scene information.

\end{itemize}

\subsection{Skeleton-Based HOI Recognition}

While RGB-based approaches dominate the HOI literature, skeleton-based methods offer lightweight and privacy-preserving alternatives:

 \begin{itemize}

\item Efficient-GCN \cite{song2022constructing}: A streamlined GCN architecture optimized for skeleton-based action recognition. However, it overlooks scene interaction nodes, limiting its application to HOI scenarios.

\item PoseC3D \cite{duan2022revisiting}: A 3D convolution-based model that processes skeletal data for action recognition. Despite its effectiveness, it lacks contextual modeling of scene information.

\end{itemize}

\subsection{Key Gaps in SOTA and Contributions}

The following limitations are observed in existing methods:

 \begin{itemize}

\item Scene Information: Most SOTA methods fail to integrate scene-specific interaction areas (e.g., fixed objects) into their learning frameworks.

\item Temporal Dependencies: Limited exploration of hybrid architectures (e.g., GCN+GRU) to model temporal dynamics in HOI scenarios.

\item Multi-Task Learning: Few approaches simultaneously classify interaction and action types, which limits their generalizability.

\item Lightweight Models: Many SOTA methods (e.g., DETR, VidTr) are computationally expensive and unsuitable for real-time applications.

\end{itemize}

The proposed method addresses these gaps by:

 \begin{itemize}

\item Embedding scene interaction nodes into the graph structure, enabling a more comprehensive representation of contextual information.

\item Introducing a hybrid GCN+GRU architecture to effectively capture temporal dependencies in human-object interactions.

\item Implementing a multi-task learning framework for simultaneous classification of interaction and action types, reducing overfitting and improving model generalization.

\item Achieving superior performance (99.25\% accuracy) on a custom HOI dataset, outperforming existing methods such as SPGCN (78.3\%) and 2G-GCN (92.4\%).

\end{itemize}

\begin{table*}[htp]
\caption{Comparison Table}\label{tab1}
\resizebox{\textwidth}{!}{%
\begin{tabular}{@{}llllll@{}}
\toprule
\textbf{Method} & \textbf{Dataset}   & \textbf{Accuracy (\%)}  & \textbf{Key Features}  & \textbf{Limitations}  \\
\midrule
DETR (2020) & HICO-DET  & 80.1 & Transformer-based, RGB data & Computationally expensive\\
Action Genome (2020)    & AG Dataset  & 85.4 &  Scene graph relationships & No skeleton-based modeling \\
Dynamic GCN (2020) & NTU RGB+D & 89.3 &  Adaptive graph connections & Lacks scene nodes, resource-heavy  \\
2G-GCN (2022)   & Custom  & 92.4 & Geometric features, GCN-based & No temporal modeling \\
SPGCN (2021)  & Custom  & 78.3  &  Scene-perception graphs & Static graphs, low accuracy\\
Efficient-GCN (2022)    & CAD-120   & 96.5 &  Lightweight, skeleton-based GCN & No scene context\\
\textbf{Proposed Method}  & Custom   &  \textbf{99.25}  & Scene nodes, GCN+GRU, multi-task learning  & Needs dynamic object handling\\
\bottomrule
\end{tabular}%
}
\end{table*}

\section{Methodology}

Recognition of Human-Object Interaction (HOI) scenarios in public places is challenging due to the complexity of the action embedding, lack of effective representation of scene information, and appropriate learning architectures, e.g., employees check-in/out by interaction with a fixed machine, etc. To achieve successful recognition of these scenarios, the scene information of related objects should be considered. Also, the relationship between human and object information should be effectively embedded within the learning model architecture. This should help to precisely identify and differentiate the human-object interaction scenario from the non-interaction one (e.g., standing while waiting for an elevator, etc.). To do so, we propose a novel HOI recognition framework that combines graph-based learning with multi-task training. The framework integrates scene interaction nodes, enabling contextual understanding, and uses a GCN+GRU hybrid architecture for spatio-temporal modeling. A multi-task learning approach jointly predicts action classes and interaction types, improving generalization. The process flow is as follows:

1. Collection of human-object interaction/non-interaction video data from real-life public places. 

2. Acquire human skeleton pose and interaction area coordinates by using an appropriate computer vision model (e.g., a pose estimation model).

3. Establish a fully connected graph representation architecture by considering action-links between human skeleton poses and interaction area nodes.

4. Select an appropriate graph learning structure using human poses and interaction area nodes, along with defining the effective sample size.

5. Maximize the recognition performance of the HOI cases by considering combined input features for training (2D human poses along with interaction area nodes) and multi-task learning approach.

\subsection{Collecting Data}

Large data collection for HOI scenarios should take place in real-world public places. This can be achieved by acquiring the licenses to collect data in specified locations. In addition, to ensure that the data are as diverse, HOI scenarios should be collected from different environments. 

Human skeleton poses and interaction areas in two dimensions (X and Y) can be acquired using tracking systems (i.e., vision or non-vision). Non-vision-based tracking approaches address social privacy problems. However, in public settings, the related collection system using sensors, such as Li-DARs, might be complex and expensive to install. To overcome this problem, vision-based systems using cameras are cost-effective, and recent related identification algorithms can disguise the faces of each individual in the scene. Figure 1 shows an example of skeleton poses acquired using Asilla product from an adult interacting with an ATM machine.

\begin{figure}[htp]
\begin{center}
    \centering
    \includegraphics[width=0.75\columnwidth]{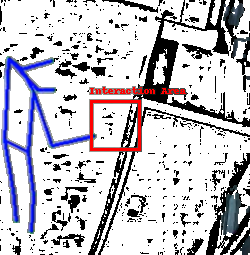}    
    \caption{Example of HOI scenario with an adult skeleton acquired using Asilla product}
\end{center}
\end{figure}

\subsection{Embedding Scene Information and Graph Representation}

Interaction areas should be decided from the collected videos by analyzing the data (e.g., heatmaps, clustering approaches, etc.). In most cases, the interaction occurs by hand. Hence, there is a need to analyze the frequent hand interaction occurrence to acquire common interaction areas around fixed objects and other possible interaction areas. 

The graph consists of human skeleton key points and scene interaction nodes. Interaction nodes represent frequently interacted regions (e.g., ATM screens), which can be identified using clustering techniques (e.g., DBSCAN) and heatmap analysis of wrist positions. Skeleton and interaction nodes should be fully connected based on human skeleton key points and wrist interaction nodes.

\subsection{Machine Learning Frameworks}

The learning structure should consider both skeleton poses and interaction area coordinates. Also, it should follow an effective, fully connected graph representation by considering action-links in between. In this context, graph network architectures have shown high performance in action recognition tasks. For instance, Graph Neural Neworks (GCNs) \cite{song2022constructing} and their combinations with Gated Recurrent Units (GRUs) \cite{xu2021scene} have shown effectiveness in the field of action recognition. Then, the learning framework should be selected and adopted from these models. 

The recommended multi-task learning framework is a hybrid GCN+GRU architecture where GCN layers represent the spatial relationships among nodes and are learned using graph convolutional layers. The node feature updates are computed as GRU Layers (temporal dependencies are captured using gated recurrent units). The proposed framework jointly optimizes primary task (action type classification) and secondary task (binary classification of interaction vs. non-interaction)

\subsection{Maximizing Recognition Performance}

To successfully recognize HOI cases, there is a need to consider combined input features for training (2D human poses along with interaction area nodes) and multi-task learning approaches. These features should be fed to the learning framework along with a fully connected graph and action links in between. In addition, to maximize the HOI recognition performance, a multi-task learning approach can be followed to learn the model, not only the action itself but also the action type, whether it envolves an interaction or not.   

\section{A Case Study of Human-Object Interaction/Non-Interaction Scenarios}

To develop an effective HOI recognition model, we considered two groups of interaction (hands on check-in/out machines, hands-on ATM machines) and non-interaction (standing, walking) classes. The studied actions mostly occur in public places (e.g., working environments, shopping malls, etc.), including normal pedestrians moving around the person who is performing the targeted action.

\subsection{Data Collection}

We collected our data from seven public cameras in shopping malls and working environments, including normal walking or standing (i.e., waiting for an elevator) (as shown in Figure 2) and interaction scenarios with fixed machines (ATM or check-in/out machines) (as shown in Figure 3). Also, different camera angles are considered so that we can ensure diverse trials as much as possible. The collected data included recorded videos from two whole different days. Then, we followed a DBSCAN clustering algorithm and heatmap analysis of right hands to identify the interaction areas (as shown in Figures 4 and 5) with respect to fixed objects in the environment (maximum of four objects). The selected interaction and non-interaction samples include a total of around twenty-seven minutes per action, while each sample size is 4 seconds (20 frames). We used the Asilla product to estimate and obtain the skeleton poses for humans in the collected videos. It includes the following 14 poses: nose, neck, ankles (right and left), shoulders (R,L), elbows (R,L), wrists (R,L), hips (R,L), and knees (R,L). Due to privacy concerns, we hide the person information and show only their skeleton poses in blue color (obtained by the Asilla pose product). 

\begin{figure}[htp]
\begin{center}
    \centering
    \includegraphics[width=1\columnwidth]{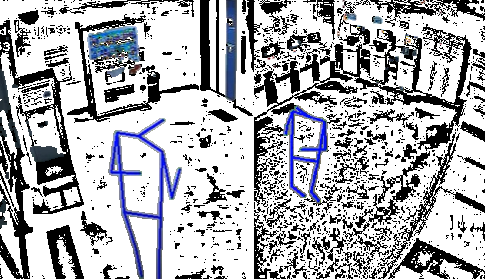}    
    \caption{Normal walking (on right) and standing (on left) from different closed environments}
\end{center}
\end{figure}

\begin{figure}[htp]
\begin{center}
    \centering
    \includegraphics[width=1\columnwidth]{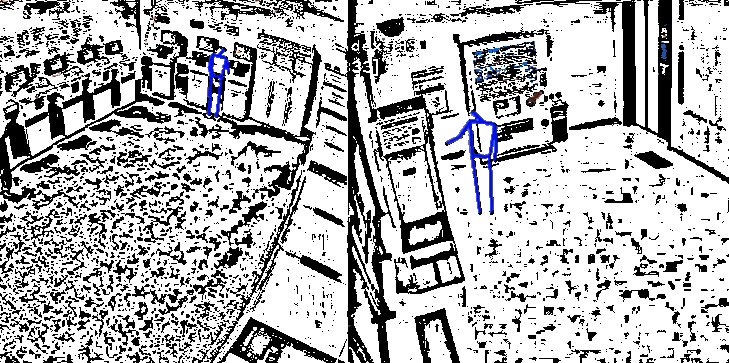}    
    \caption{Interaction scenario with ATM machine (on right) and check-in/out machine (on left) from different closed environments}
\end{center}
\end{figure}

\begin{figure*}[htp]
\begin{center}
    \centering
    \includegraphics[width=2.25\columnwidth]{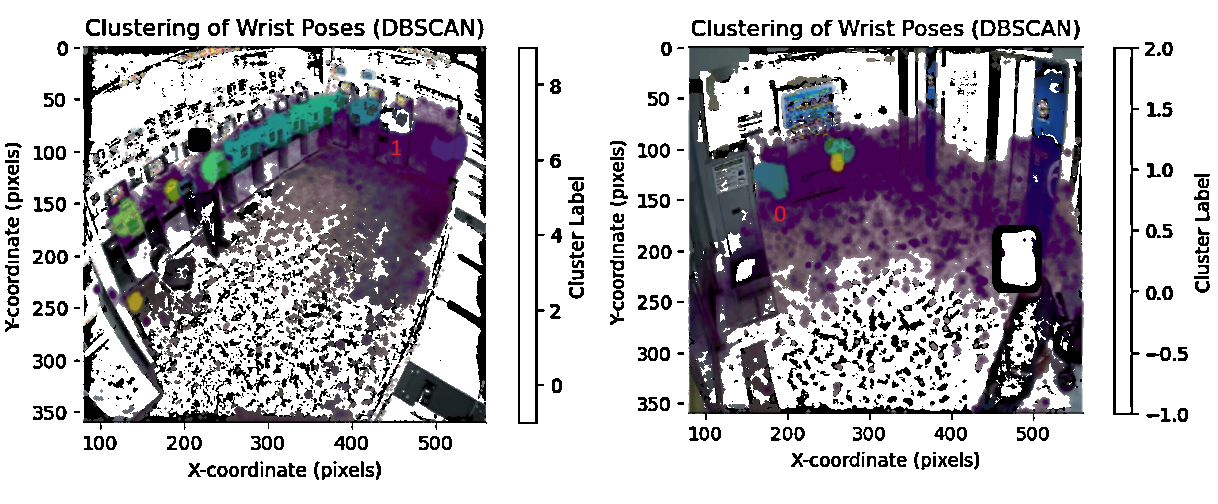}    
    \caption{Right-hand DBSCAN clustering algorithm results with example of interaction areas numbered in red (0 and 1) from two different environments}
\end{center}
\end{figure*}

\begin{figure*}[htp]
\begin{center}
    \centering
    \includegraphics[width=2\columnwidth]{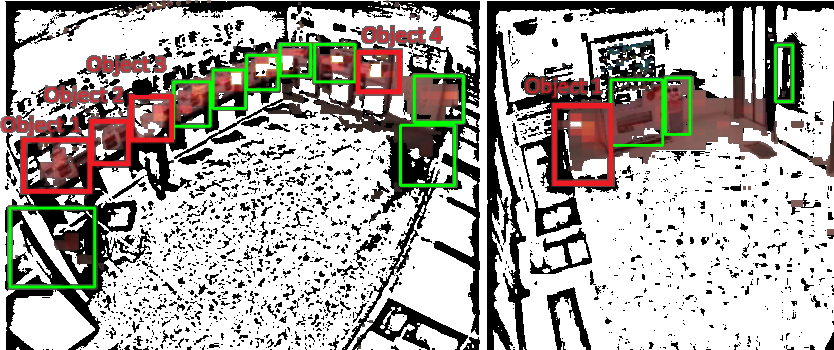}    
    \caption{Right-hand heatmap analysis with highlighted interaction areas (in red boxes) from two different environments}
\end{center}
\end{figure*}

\section{Results}

We include in this section the training results following the Efficient-GCN \cite{song2022constructing}, GCN+GRU, and SPGCN \cite{xu2021scene} training structures (explained in the next subsection). Based on our several training trials using different sample sizes, we found that 4 seconds (20 frames) is the most effective duration to obtain the highest recognition performance of our studied actions. Based on that, a series of four-second skeleton poses (14 poses per window) plus at most four interaction areas (8 nodes per window) are fed to the network. Where each interaction area is represented by upper left and lower right nodes. In cases of non-existing interaction areas, we consider these values as zero. We also followed a fully connected graph representation between skeleton poses and interaction area nodes (skeleton bones and action-links). Finally, we normalized the data in the two dimensions (X, Y) based on the data range of (640, 360) pixels and processed the data (400 samples per action) into 80\% for training, 10\% for validation, and 10\% for testing.  

\subsection{Training Frameworks}

To evaluate the proposed methodology, we experimented with two updated training frameworks: GCN+GRU and SPGCN. The GCN+GRU architecture extends the basic GCN structure \cite{song2022constructing} by incorporating Gated Recurrent Units (GRUs) to capture temporal dependencies more effectively. Specifically, GRU layers were added after each GCN block, followed by Global Pooling and a Softmax layer, aligning the structure with the SPGCN design. The SPGCN framework \cite{xu2021scene}, on the other hand, was configured with three hierarchical GCN blocks, each tailored to process increasing levels of abstraction. The first block had 64 output channels, stride 1, and depth 3; the second block expanded to 128 output channels with stride 2; and the final block utilized 256 output channels, also with stride 2. To enhance the temporal learning capabilities, a three-layer GRU block with linear layers was incorporated, followed by batch normalization, dropout layers, and a fully connected output layer. These frameworks were rigorously tested to determine the optimal configuration for recognizing human-object interactions and non-interaction scenarios.

The training is performed using Python and PyTorch, where a series of (X, Y) human skeleton poses plus interaction area nodes (uper left and lower right) are fed to the network with scaled samples of four seconds (20 frames per window) as a single window, where each window consists of separate interaction or non-interaction actions (4 classes of standing (1), walking (2), hands-on check-in/out machines (3), and hands-on ATM machines (4)). We also followed a multi-task learning approach where additional group labels (Non-interaction (0): standing, walking, Interaction (1): hands-on check-in/out machines, hands-on ticketing machines) are provided to the training structure. The size of the data used for training is balanced (200 samples per action).

\subsection{Baseline (Efficient-GCN)}

The baseline is trained using only 2D skeleton poses versus combined inputs with interaction area nodes and performing the training with single or multi-CrossEntropy Loss functions. This baseline is to be compared with the updated Efficient-GCN structures, which are shown in the following subsection. The resultant testing accuracy has shown an improvement from 96.5\% to 99\% with 50 epochs.

\subsection{Efficient-GCN Vs. Updated Architetures}

To optimize human-object interaction recognition, we considered training using combined input features of 2D skeleton poses and interaction area nodes with single and multi-loss functions using updated Efficient-GCN structures (GCN+GRU and SPGCN). The results in Table 1 show that most of the improvements are related to the use of object interaction area information, while the effect of multi-task learning is not clear. Also, the GCN+GRU structure showed some improvements over epochs (10–50) if compared with the Efficient-GCN structure. The best achieved performance is 99.25\% which outperforms the baseline with 2.75\%.

\begin{table*}[ht]
\centering
\caption{GCN, GCN+GRU, and SPGCN Test Results}\label{tab1}%
\begin{tabular}{|l|l|l|l|l|l|l|l|}
\hline
\multirow{3}{*}{} & \multirow{3}{*}{} & \multicolumn{2}{|c|}{Num Epochs 10} & \multicolumn{2}{|c|}{Num Epochs 20} & \multicolumn{2}{|c|}{Num Epochs 50} \\
\cline{3-8}
 &  & Without Obj. & With Obj. & Without Obj. & With Obj. & Without Obj. & With Obj. \\
\hline
\multirow{2}{*}{GCN} & Single-Loss & \textbf{85\%} & 74.75\% & 98\% & \textbf{98.25\%} & 96.50\% & \textbf{99.25\%} \\
 & Multi-Loss & \textbf{92\%} & 88.50\% & 96.25\% & \textbf{99\%} & 96.75\% & \textbf{99\%} \\
\hline
\multirow{2}{*}{GCN+GRU} & Single-Loss & \textbf{84.25\%} & 81.75\% & 97.50\% & \textbf{98.75\%} & 97.75\% & \textbf{99.25\%} \\
 & Multi-Loss & \textbf{89\%} & 81.50\% & 96.50\% & \textbf{98.50\%} & 97.25\% & \textbf{99.25\%} \\
\hline
\multirow{2}{*}{SPGCN} & Single-Loss & \textbf{81.25\%} & 80.75\% & 96.25\% & \textbf{98\%} & 96.25\% & \textbf{98\%} \\
 & Multi-Loss & \textbf{84.50\%} & 80.25\% & 96\% & \textbf{98\%} & 96.75\% & \textbf{99\%} \\
\hline
\end{tabular}
\end{table*}

The confusion matrices of the (GCN+GRU) architecture's best results with single and multi-loss training functions are shown in Figures 6 and 8. Also, the training and evaluation loss curves are shown in Figures 7 and 9, respectively. The lowest number of failure cases (3 samples) are shown from the models, which consider interaction nodes as additional input features. Single and multi-loss training best performance is almost the same with a slight difference in the type of failure cases (interaction class (IC)/non-interaction class (NIC)). However, the multi-task learning might have improved the loss over epochs slightly.

\begin{figure*}[htp]
\centering
\includegraphics[width=1\textwidth]{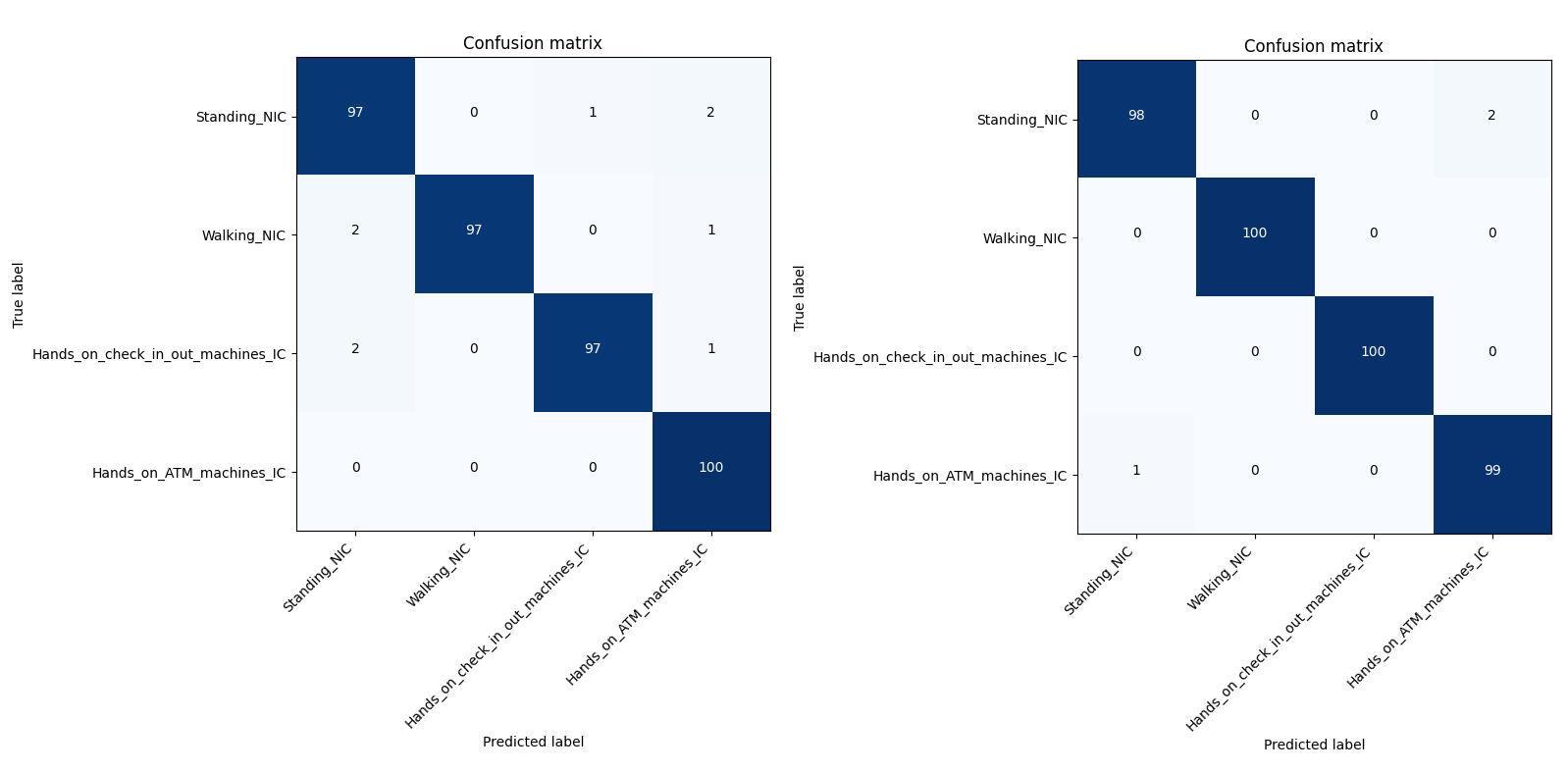}
\caption{(GCN+GRU) Without interaction nodes (on left) and with interaction nodes (on right) Confusion matrices on test data}
\label{fig:confusion_matrix}
\end{figure*}

\begin{figure*}[htp]
\centering
\includegraphics[width=1\textwidth]{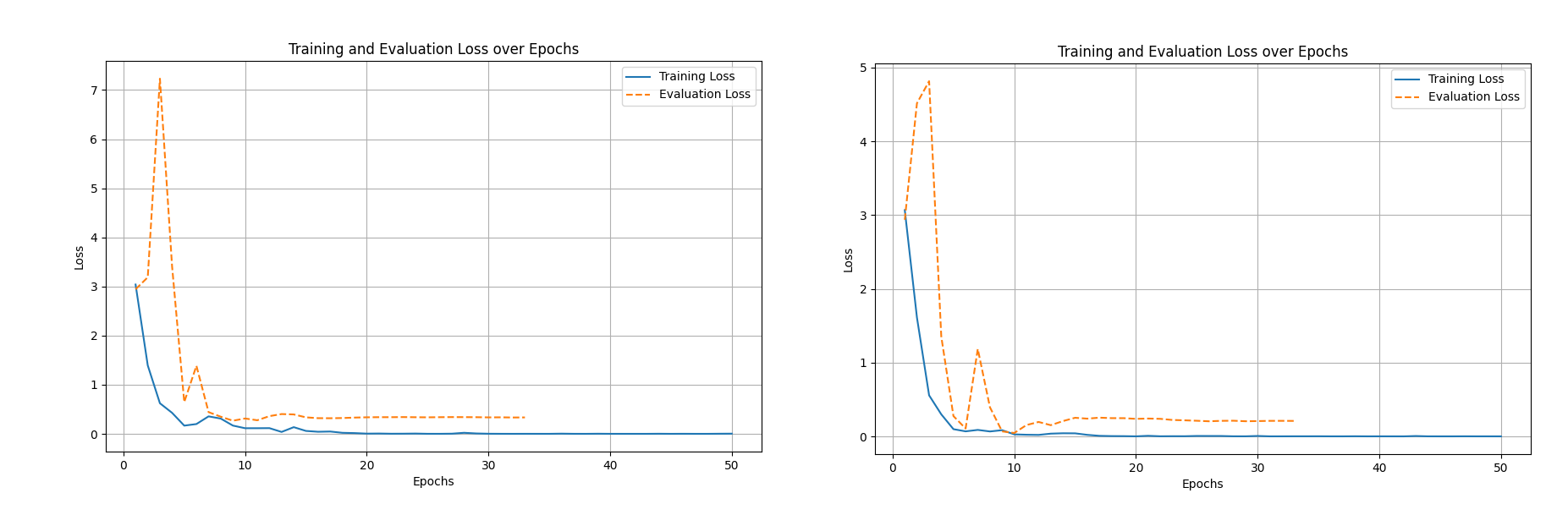}
\caption{(GCN+GRU) Without interaction nodes (on left) and with interaction nodes (on right) single-loss training and evaluation results}
\label{fig:loss}
\end{figure*}

\begin{figure*}[htp]
\centering
\includegraphics[width=1\textwidth] {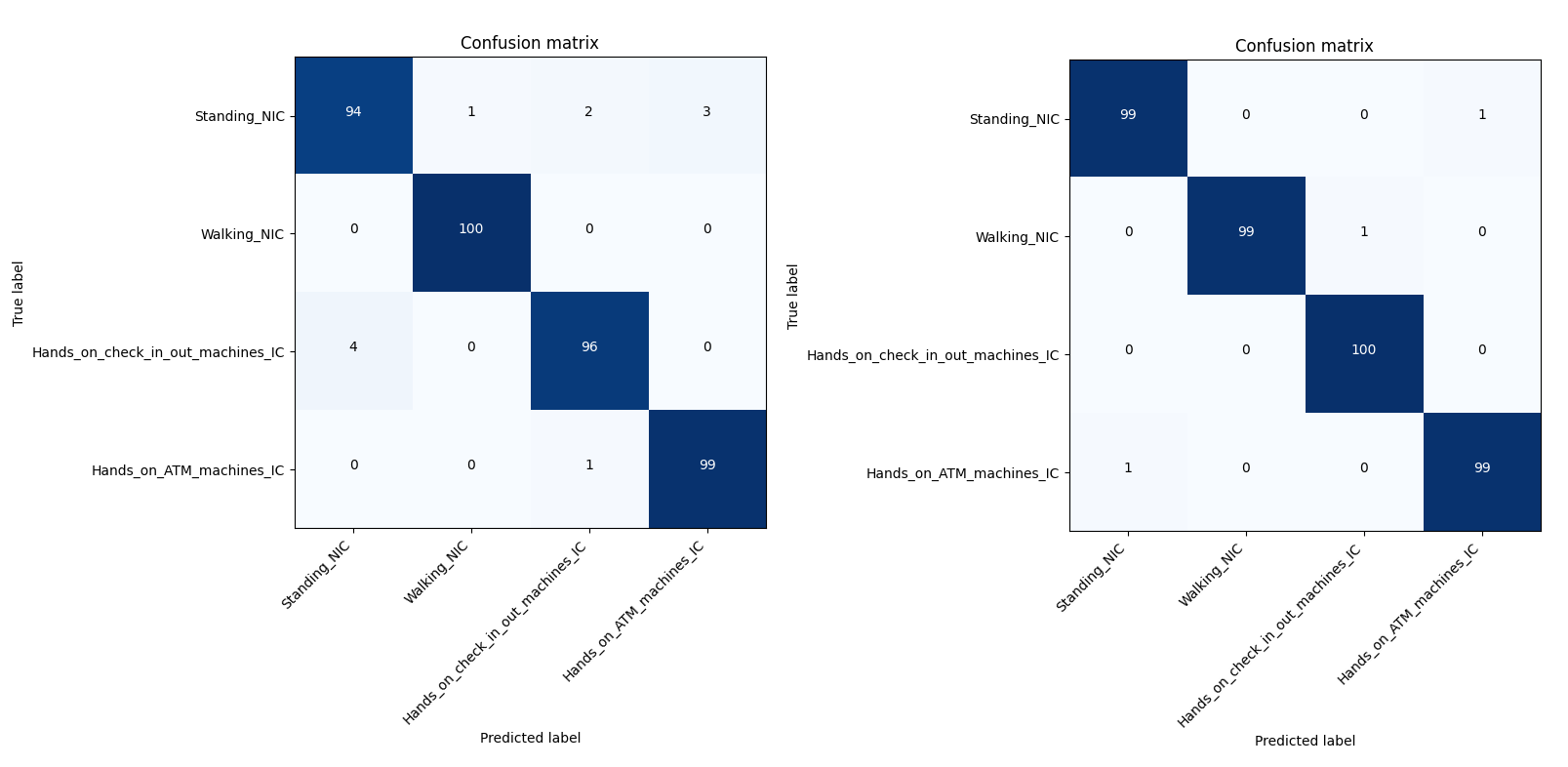}
\caption{Multi-Task Learning (GCN+GRU) Without interaction nodes (on left) and with interaction nodes (on right) Confusion matrices on test data}
\label{fig:confusion_matrix}
\end{figure*}

\begin{figure*}[htp]
\centering
\includegraphics[width=1\textwidth]
{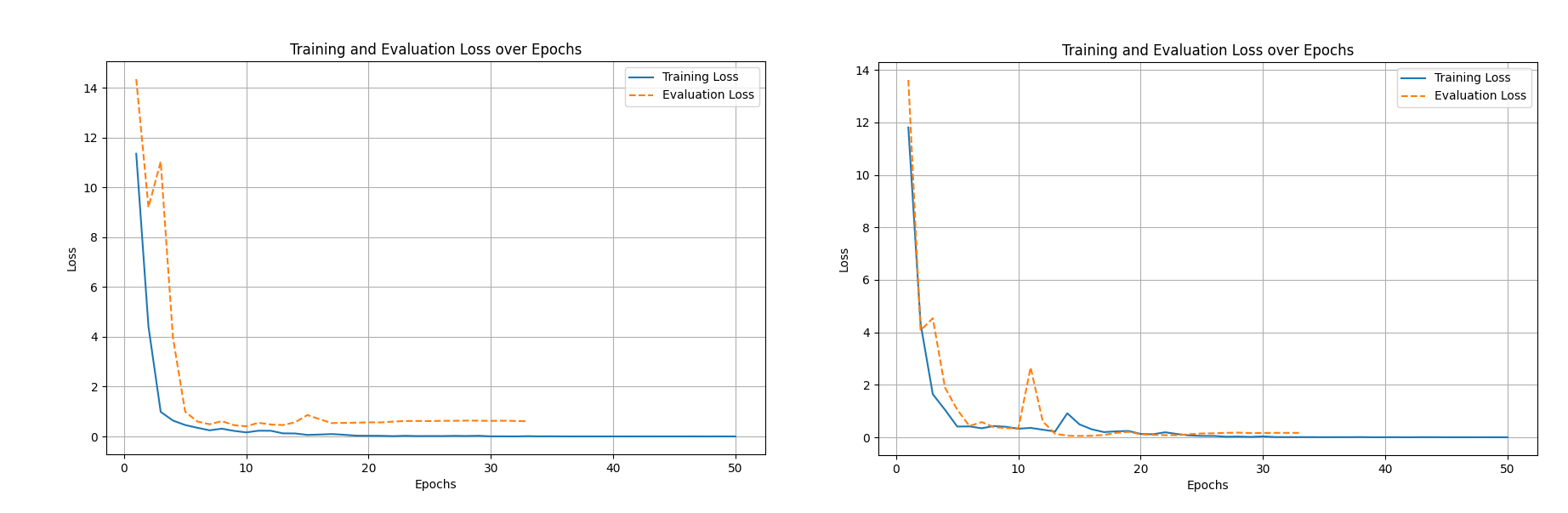}
\caption{(GCN+GRU) Without interaction nodes (on left) and with interaction nodes (on right) multi-loss training and evaluation results}
\label{fig:loss}
\end{figure*}
\nopagebreak
\vspace{-1em} 

\section{Discussions}

Our results highlight the significant role of scene information in improving human-object interaction and non-interaction action recognition. By embedding interaction area nodes into the graph learning architecture, the model successfully captured nuanced contextual details, which enhanced its ability to distinguish between subtle action types. Although the inclusion of a multi-task learning framework did not yield a clear or consistent effect, it showed a marginal improvement in loss reduction over successive epochs, suggesting its potential for further refinement in future studies. Additionally, incorporating GRU layers after GCN blocks (as in the GCN+GRU framework) demonstrated noticeable performance gains by better modeling temporal dependencies. However, the introduction of additional GCN blocks or modifications in the Efficient-GCN structure, such as in the SPGCN framework, introduced architectural complexity that did not necessarily translate to improved results. These findings underscore the importance of balancing model complexity with performance enhancements in designing action recognition systems.

Many existing SOTA methods for human-object interaction recognition, such as SPGCN \cite{xu2021scene} and 2G-GCN \cite{qiao2022geometric}, rely heavily on geometric or spatial features but fail to incorporate contextual information from the scene. Furthermore, they lack mechanisms to model temporal dependencies effectively, which are crucial for capturing the dynamics of HOI scenarios. Transformer-based methods such as DETR have shown promise for general object detection but have yet to demonstrate consistent performance for skeleton-based HOI recognition.

By integrating scene interaction nodes into a graph-based learning architecture, our method outperforms SOTA methods by achieving a recognition accuracy of 99.25\%, compared to SPGCN (78.3\%) and 2G-GCN (92.4\%). Additionally, our use of GRU layers after GCN blocks enables better temporal modeling, which is critical for recognizing nuanced interactions over time. While transformer-based methods focus on RGB frames, our skeleton-based approach provides a lightweight alternative for real-time applications.

Our main contributions are summarized into the following points:

 \begin{itemize}
 
\item Contextual Modeling: Scene interaction nodes enhance contextual understanding, reducing misclassifications in ambiguous scenarios.

\item Temporal Dynamics: The GCN+GRU hybrid architecture effectively models spatio-temporal relationships.

\end{itemize}

While our method significantly outperforms existing approaches, several limitations remain:

 \begin{itemize}

\item Dataset Generalization: The current evaluation is conducted on a custom dataset. Future work will validate the model on public benchmarks such as HICO-DET, AVA, and CAD-120.

\item Dynamic Object Interactions: The model is limited to fixed-object interactions. Extending the framework to dynamic interactions (e.g., moving objects) will require integrating object tracking algorithms.

\end{itemize}

Future research will focus on addressing these limitations by extending the model to handle dynamic object interactions, experimenting with larger and more diverse datasets, and optimizing the multi-task learning framework for real-time applications.

\section{Conclusions}

We propose a new skeleton-based action recognition approach to improve human-object interaction action recognition using scene information and a multi-task learning approach. To evaluate the proposed method, we collected real data from seven public environments and prepared our data set, which includes interaction classes of hands-on fixed objects (ATM ticketing machines, check-in/out machines) and non-interaction classes of walking and standing. Most of the improvements are related to the use of object interaction area information, while the effect of multi-task learning is not clear. Also, the (GCN+GRU) structure showed some improvements over epochs if compared with the Efficient-GCN structure. The best achieved performance is 99.25\% which outperforms the baseline with 2.75\%.

\section*{Acknowledgment}

The technical support provided by Asilla Vietnam team members in the pose estimation algorithm is highly appreciated. Also, special thanks to Mr. Trung Tran Quang for the fruitful discussions and comments during the data preparation process.

\section*{Availability of data and material}

The datasets generated during and/or analyzed during the current study are available from the corresponding author on reasonable request.

\section*{Conflict of Interest}

The authors declare that they have no conflicts of interest.

\end{document}